# Predicting delays in Indian lower courts using AutoML and Decision Forests

Mohit Bhatnagar[1] and Shivraj Huchhanavar[2]

[1] Jindal Global Business School, OP Jindal Global University, Sonipat, India
[2] Jindal Global Law School, OP Jindal Global University, Sonipat, India
(mohit.bhatnagar, shivaraj.huchhanavar)@jgu.edu.in

**Abstract** This paper presents a classification model that predicts delays in Indian lower courts based on case information available at filing. The model is built on a dataset of 4.2 million court cases filed in 2010 and their outcomes over a 10-year period. The data set is drawn from 7000+ lower courts in India. The authors employed AutoML to develop a multi-class classification model over all periods of pendency and then used binary decision forest classifiers to improve predictive accuracy for the classification of delays. The best model achieved an accuracy of 81.4%, and the precision, recall, and F1 were found to be 0.81. The study demonstrates the feasibility of AI models for predicting delays in Indian courts, based on relevant data points such as jurisdiction, court, judge, subject, and the parties involved. The paper also discusses the results in light of relevant literature and suggests areas for improvement and future research. The authors have made the dataset and Python code files used for the analysis available for further research in the crucial and contemporary field of Indian judicial reform.

## 1. Introduction

The expense, delay, and pendency have been enduring challenges facing the Indian courts [1]. Studies have underlined various systemic inadequacies that affect the celerity, cost-effectiveness, and administrative efficiency of courts in India [1]. Technology is seen as a panacea in optimizing judicial processes and case management. The judiciary, since 2005, has worked to build a robust computing infrastructure in courts across India [2]. To facilitate data-driven case and court management, the judiciary has also developed the National Judicial Data Grid [3]. However, the problem of pendency has not eased; on the contrary, pendency has increased significantly in recent years [4].

Against this backdrop, this paper demonstrates the potential for an AI model to predict the delay of court cases, using information such as the type of cases, and names of judges, advocates and parties. This predictive analysis would not only help litigants, advocates, and judges make informed decisions, but it would also help identify systemic issues affecting the efficiency of courts and judges. This research is, to some extent, a departure from research that aims to predict or qualitatively assess judicial outcomes. Although the latter applications of machine learning and AI are no less relevant; however, India's judiciary faces elementary challenges that can only be addressed by improving the efficiency of judicial actors and institutions, and predictive modelling of the duration and delays of cases would be a step forward in this direction.

## 2. Setting the context

AI and machine learning could strengthen judicial administration in various ways; the range of applications can encompass intelligent analytics, research and computational tools, with the eventual goal of implementing predictive justice [5]. These tools have the potential to provide comprehensive legal briefs on cases, including legal research and identifying crucial points of law and facts; thus, expediting the judicial process. Predictive AI tools can complement human judgment in the adjudicative process [6]. AI tools can also enhance the administrative efficiency of the courts and judges. It can facilitate e-filing, filtering, listing, notifying and prioritization of cases [7]. Additionally, intelligent tools, such as legal bots, can be designed to assist (potential) litigants in making informed decisions regarding their legal rights and to provide cost-effective access to basic legal services. Likewise, as this paper demonstrates, AI can also help predict the delays for a case in court.

For the purpose of this study, we see extant literature into two categories. Firstly, literature that endeavours to predict court case outcomes through artificial intelligence, and secondly, literature highlighting the effectiveness of AI in predicting court case delays. Over the past two decades, scholars have made significant attempts, resulting in noteworthy successes, to predict the outcomes of different types of court cases in several jurisdictions, notably China [8], the United States [9], the European Union [10], France [11], the United Kingdom, Brazil [12] and India [13].

It is pertinent to note that predictive AI models have been employed mostly for outcome identification, outcome-based judgement categorisation, and outcome forecasting [14]. Identification of verdict in a full-text of a judgement is called outcome identification; whereas judgement categorisation is the task of categorising documents, based on the outcome. Likewise, as the term suggests, outcome forecasting involves the prediction of future decisions of courts [15]. However, so far, there has been no emphasis on predicting pendency in courts, this is true for India and elsewhere.

The Indian judiciary is keen on adopting AI tools in judicial administration. The judiciary has demonstrated foresight by being an early adopter of technology and making progress, albeit limited, in exploiting the capabilities of AI and data-driven insights. Advancements in ICT-enabled judicial administration are attained through the eCourts Mission Mode Project; the project set up a foundation for e-courts, equipping courts with fundamental computing hardware. In recent years, the Indian judiciary has taken a notable step to harness the potential of AI. In 2019, the then Chief Justice of India, Justice Bobde, launched the beta version of a neural translation tool called SUVAS [16]. This software can translate judicial rulings in English to nine vernacular scripts, and vice-versa. In 2021, the Supreme Court of India launched an AI portal: SUPACE, which aims to utilize machine learning to manage large amounts of case data [17]. The Court has a dedicated Artificial Intelligence Committee, headed by a Supreme Court judge, to explore the use of AI in judicial administration [18].

In addition, with the support of the government, the judiciary has built the National Judicial Data Grid (NJDG). The NJDG is a database of orders, judgments and case details of 18,735 District & Subordinate Courts and High Courts. Data is updated on a near real-time basis by the connected district and taluka courts. It provides data relating to judicial proceedings and decisions of all computerized district and subordinate courts in India. The relevant data of all High Courts are available on the Grid [19]. Litigants may access case status, pendency data and orders of courts using the Grid. The Grid has the ability to perform drill-down analysis based on the age of the case as well as the State and District; it provides court-wise, subjective-wise and age-wise pendency of subordinate and high courts [19]. The NJDG provides the Central and State Governments with exclusive access to data using a departmental ID and access key. This facilitates institutional litigants' access to the NJDG data for monitoring and evaluation of cases where such institutions are one of the parties to pending litigations [19]. However, the Grid cannot produce predictive analyses of the case pendency and public presentation of data by NJDG is currently limited to descriptive analysis.

Despite the ongoing efforts to digitize court judgments, particularly in lower courts, many of these judgments remain in paper form. Moreover, many subordinate courts lack basic computing infrastructure, thereby causing inefficiencies in the system. The current advancements in the application of AI in India's judicial administration are overlooking the requirements of the lower courts. The lower courts are the real face of India's judiciary as they deal with 87.4% of all pending cases in the country [20]. Besides the burden of backlog, the subordinate courts face acute shortages of judicial personnel, infrastructure and funding [21]. The challenges facing the lower courts further emphasise the need for innovative and frugal solutions; AI and machine learning can address some of the challenges facing the subordinate judiciary in India by identifying the major pain points and roadblocks. It is in this background, the current study works on delay and pendency prediction with data of subordinate courts. As NJDG holds categorised data on geographical, subject-wise, court-wise, judge-wise and purpose-wise data holds great promise for predictive insights and also in discovering the relevant aspects that cause the delays. Therefore, building a predictive model addressing the delay or their causes for cases clogged in the judicial system or analysing the importance of aspects that cause the delay is a logical next step.

In this study, to explore the feasibility of predictive insights, we train a multi-class classification model using Google's Vertex AI and AutoML framework on a publicly available dataset of 4.2 million cases. The AutoML framework also identifies the feature importance matrix that we analyse for the causes of the delay. We also train a binary classification model that can more accurately predict delays for cases which take more than the average duration of delays which are found to be around 3 years in the dataset. The overall methodology followed is depicted in Fig.1.

**Fig.1**. Research Methodology Workflow

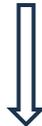

| Process | Output | Tools |
|---|---|---|
| • Data pre-processing<br>• Feature Engineering<br>• Multi-classification modelling<br>• Binary-classification modelling<br>• Model evaluation & testing | • Merged CSV file<br>• Feature vectors<br>• Trained Classification Model<br>• Trained Classification Model<br>• Classification results & feature importance | • Big Query (Google)<br>• Custom python code<br>• Vertex AI AutoML (Google)<br>• Custom Python code, scikit-learn<br>• Vertex AI AutoML, scikit-learn |

## 3. Methodology

### 3.1 Dataset & Pre-processing

The dataset we use is taken from the work of Bhowmick and others [22]. The dataset was constructed from case records scraped from the e-courts platform (https://ecourts.gov.in/) maintained by the Indian government and includes cases from 2010 to 2018. The dataset covers data from 7,000+ district and subordinate trial courts, which are staffed by over 20,000 judges. The dataset is made available by the authors for download at https://www. devdatalab.org/judicial-data and contains data of over 77 million cases. Data about these cases are tracked and updated till 2020, thereby providing comprehensive coverage over the period from 2010 to 2020. There is no personally identifiable information in the dataset. The metadata descriptions for the different features are also provided by the authors.

The dataset is organized into multiple year-wise files from 2010 to 2018 in CSV and DTA format. The files include the filing dates, the first, last and next dates of hearing for the case. In cases where the judgment has been delivered the date of the decision is provided. The state, district and court details are included along with the gender details of the petitioner, defendant, and their respective lawyers. Other important data includes the position of the presiding judge, the type and the purpose of the case. Separate metadata files are also included to augment the information available in the year-wise data file. Data for the judges' gender, tenure and the applicable Act and Section for criminal offences is separately made available with a unique case Id acting as the index to the data in these files. Certain features like the date of the first, next and last listing, purpose name and disposition were dropped from the analysis as they would ne be available on the filling date. Table 1 provides a summary of the features that were included in the analysis.

**Table 1.** Feature Description

| Column Name | Description | Usage in Model |
|---|---|---|
| date_of_filing | Date when the case was filed | The target variable is calculated based on the date of decision - the date of filing |
| date_of_decision | the case reached a decision Date when | |
| state_code | Unique state code | Categorical Input features for Geographical Identifiers are calculated as the combinations |
| dist_code | Unique District Code for every district in a state | |
| court_no | Unique Court number for every court in a district | |
| judge_position | The judge's position as per the existing hierarchy | |
| female_judge_filing | Gender for the judge under whom the case was filed | |
| female_judge_decision | Gender of the judge by whom the decision was delivered | |
| female_adv_pet | Gender of the advocate from the petitioner | |
| female_adv_def | Gender of the advocate from the defendant | |
| female_petitioner | Gender of the petitioner | |
| female_defendant | Gender of the defendant | Categorical Input feature related to the case characteristics |
| type_name | The type of case | |
| section | Section under which case is filed | |
| act | Act under which the case is filed | |
| criminal | Whether this was a criminal case | |
| number_sections_ipc | Number of sections charged in criminal cases | |
| bailable_ipc | Flag to indicate if it is bailable | |

The overall dataset is of 77 million cases comprising cases filed from 2010 to 2018 and decided till 2020. In the current study, however, we include 4.2 million cases which are filed in the year 2010 and for which data has been captured for their outcomes till 2020. This approach ensures we include cases with the longest evaluation period from the dataset. Including cases filed from 2011 to 2018 in our model will bias the model towards cases which have been resolved early. Since the data is spread over multiple files and the files were large in size (over 1 GB) it is difficult for desktop applications like MS Excel to easily process them for data manipulation. We leveraged the Big Data warehouse Big Query from Google to merge all data into a single CSV file for the year 2010 for analysis. The total number of cases was 4.19 million.

**3.2 Feature Engineering**

The transformation of target and input features was required for converting the date fields to durations and subsequently into categories. The target (or the predicted) variable in the AutoML model was defined as one of the categories of pendency for a judgement, namely "< 1 year", "1 up to 3 years", ">3 up to 5 years", ">5 up to 10 years" and "> than 10 years". The class boundaries were chosen based on the categories available in the e-courts platform where the dataset was collected. Ongoing cases, where decision dates were not available, were marked in the last category.

In the custom binary classification model the model predicted one of the two categories of "<than 3 years" and "3 years and >". The 3-year boundary was chosen for the binary classification model as it approximated the boundary for predicting cases with more than average delays. The average duration of delays from filling to decision for all completed cases in our dataset was found to be 36.2 months with the median value being 31 months.

Data points were dropped if there was any discrepancy in the timelines (e.g. if the filing date was older than the decision date). Cases with filing, first hearing and decision dates earlier than 2010 were also dropped from further analysis. The categories for the target variable and feature variables were obtained by calculating the intermediate period in months or days between the dates and these were used further. A few significant data variables that were dropped include judge-related data like their tenure and if during the case there was any change of the judges. The same was done for 2010 data which had a significantly large number of missing values. Missing values were imputed with a "Not Available" string token.

### 3.3. Data Split

Post-cleaning, 3.62 million (86.4% of the overall) data points remained and the standard 80:20 split for training and validation/test split was followed for model building. The AutoML model further split the validation dataset into 10% validation and 10% test data. The total number of data points available for training was 2.90.M and 0.72M for validation for the custom model. No hyper-parameter tuning was employed for the custom models and hence a three-way split as done for the AutoML model was not required. Stratifying was used during the train test split in the custom model development to ensure a proportionate representation of all target classes. Google Vertex AI, AutoML model provides out-of-the-box functionality for ensuring stratification for the train test split.

## 4. Modelling and Results

AutoML is a rapidly growing field in the domain of machine learning that aims to automate the process of building machine learning models [23]. It involves the use of algorithms and techniques to automate tasks such as feature selection, hyperparameter tuning, and model selection. Practitioners commonly use AutoML for initial prototyping to check the quality of the data and perform feature selection [24]. This usage of AutoML helps quickly identify the most relevant features and obtain a quick preliminary understanding of the relationships between the features (independent variables) and the target (dependent variable).

Once the initial validation and prototyping have been done, it is common to build custom models [25]. Decision forests and in particular, Gradient Boosting Decision Trees (GBDT) are popular choices given that they have produced state-of-the-art results in several online competitions on sites like Kaggle. The Decision forests are also easier to train and interpret [26]. Further, for developing the custom models, binarization techniques are employed where multiple binary classifiers are used to model the multi-classification problem [27]. This approach of using multiple binary classification models can provide a more nuanced understanding of the problems. For example, if a multi-class classification model is used to predict the pendency of cases, it may not be possible to distinguish between cases that are highly likely to be resolved soon and cases that are likely to be pending for a long time. By using multiple binary models, each model can be trained to handle a different part of the prediction, allowing for a more nuanced understanding of the problem.

In our model-building process, we utilize the aforementioned methodology. First, we utilize Google's Vertex API framework to construct an AutoML multi-class classification model, which predicts pendency into five classes. Next, we use the identified features to build a custom binary classification model in Python only focusing on a binary classification of the delayed cases into greater than or less than 3 years classes. This model employs scikit-learn's decision tree and ensemble classifiers of Bagging, Random Forest, and XGBoost.

Given the need for and importance of Explainable AI (XAI) for interpretable and actionable insights in domains which impact diverse groups of people we use the SHAP [28] framework. The SHAP framework provides value plots which are a way to measure the contribution of each feature to complex prediction models. For our study, the plots can be useful in determining the most important aspects of pendency.

For the binary classification model, the feature importance matrix is readily available in the scikit-learn decision tree classifiers. However, for obtaining the feature important, Label encoders were used for the transformation of input categorical features. Label Encoders since they put an artificial order to the categorical variables are not the best-performing encoders [28], however, their use makes the feature importance matrix readily available. The interpretation of the feature importance matrix for the models facilitates the prioritization of areas where the focus needs to be provided for improving the court processes for pendency reduction.

### 4.1. AutoML model

**Model Training** Google AutoML Vertex AI automates the process of finding the best model and tuning its hyperparameters to achieve the best performance. The training time was 9 hours and 20 minutes which included 300 tuning trials with different Neural Networks and Boosted Tree Models with different configurations of layers and trees respectively. The best-performing model which was proposed by AutoML was an ensemble of 25 neural network models with 2,3 or 5 hidden layers with embedding sizes of 16, 128 or 1024. Dropout values used in the different models were 0.25, 0.375 or 0.5.

**Results** The results showed varying levels of accuracy for the different classes. The model was best performing in "< than 1 year" and "5 to 10 years" with an accuracy of 83% and 72% respectively, with F1 scores of 0.66 and 0.54, PR AUC of 0.77 and 0.66, ROC AUC of 0.90 and 0.85, Precision of 73.2% and 69.7% and Recall of 60.7% and 43.8%. Overall, the model had an F1 score of 0.47, PR AUC of 0.60, ROC AUC of 0.86, Precision of 62.9% and Recall of 37.3%.

The SHAP feature importance metrics indicated that the geographic identifiers namely the specific courts, specific districts and the state contributed almost 50% while the type of case, section and the judge position were the other important features. The detailed results are presented further in Fig. 2 (feature importance), Table 2 (confusion metrics) and Table 3 (accuracy metrics). The findings indicated that the model performance was good in classifying cases only in <1 year and 5 to 10 years band, but the performance was found inadequate in all other classes. The feature importance metrics indicated that the shortlisted features (16 in number) all contributed to the model prediction.

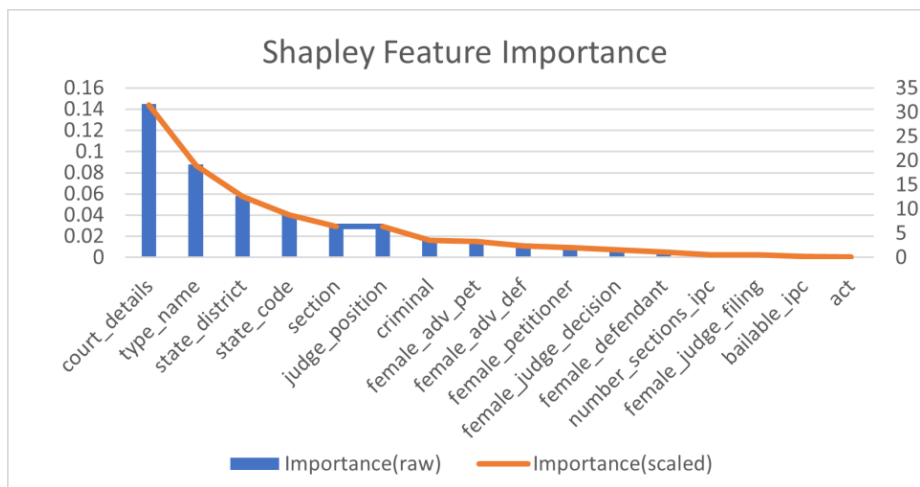

**Fig. 2.** Feature Importance (AutoML)

**Table 2.** Confusion Matrix (AutoML)

| True /Predicted label | < 1 year | 1 upto 3 years | 3 upto 5 years | 5 upto 10 years | >10 years |
|---|---|---|---|---|---|
| < 1 year | **83%** | 16% | 0% | 0% | 0% |
| 1 to 3 years | 27% | **55%** | 13% | 6% | 0% |
| 3 to 5 years | 33% | 8% | **27%** | 33% | 0% |
| 5 to 10 years | 10% | 3% | 16% | **72%** | 0% |
| >10 years | 5% | 0% | 2% | 93% | 0% |

**Table 3.** Accuracy Matrix (AutoML)

|  | All Labels | < 1 Year | 1 upto 3 Years | > 3 upto 5 Years | > 5 upto 10 Years | > 10 years |
|---|---|---|---|---|---|---|
| **PR AUC** | 0.60 | 0.76 | 0.51 | 0.36 | 0.66 | 0.02 |
| **ROC AUC** | 0.86 | 0.90 | 0.72 | 0.74 | 0.85 | 0.58 |
| **Log loss** | 0.99 | 0.34 | 0.55 | 0.47 | 0.38 | 0.001 |
| **F1 score** | 0.47 | 0.66 | 0.39 | 0.19 | 0.54 | 0.07 |
| **Precision** | 62.9% | 73.2% | 54.7% | 37.6% | 69.7% | 100% |
| **Recall** | 37.3% | 60.7% | 29.8% | 12.4% | 43.8% | 4.5% |

## 4.2 Custom Model

**Model Training** The custom model was developed in Python using the Google Colab Pro development environment. The scikit-learn, Python and machine learning packages were used for the modelling. Google Colab Pro was used as the RAM requirements were larger than those provided by the free version of Google Colab and consequently lead to code crashes. The code file, along with the cleaned dataset is made available at github at https://github.com/mb7419/pendencyprediction, for easy reproducibility of the results. The custom model included all the features included in the AutoML model since all showed some contributions in the feature importance. As discussed earlier the target variable was collapsed into two classes, one with "< than 3 years" and the other one was "3 years and >".

The decision tree classifier was used as a baseline, with a max tree depth of 10. The tree depth was set to avoid overtraining. Further the ensemble classifiers of Random Forest, Bagging and XGBoost were used to improve on the baseline model. The binary classification metrics, equivalent to the AutoML model, were used for evaluation. Label encoder was used for transforming the categorical features in these models as it is difficult to interpret when one-hot-encoding is used.

The feature importance matrix was determined based on the classifier that gave the best-performing model. The classifier that gave the highest accuracy and performance was also used to train a separate model using scikit-learns one-hot-encoding. TruncatedSVD was further used for dimensionality reduction with the top 200 components used for modelling. This model was developed to get improved or equivalent performance to the label encoder-based categorical feature encoding.

**Results** Random Forest achieved the best accuracy scores among the four different classifiers used. An accuracy of 81.5%, and the weighted average precision, recall and F1 were found to be 0.82, PR AUC was 0.40 and ROC AUC of 0.81. The confusion matrix for the results is presented in Fig. 3. The model built by using One-hot-encoding of the categorical variables improved the accuracy to 81.7% and the performance of the model was almost similar. The Bagging classifier was a close second in classification performance and the XGBoost out-of-the-box also provided similar performance. We have not conducted further hyperparameter tuning as the performance of the Random Forest classifier was found adequate for our use case and validated the hypothesis that it was possible to improve the performance of the AutoML model with a custom binary classification model. The comparative result with all classifiers is available in Table 4 and the Feature Importance is available in Fig. 4. The overall results indicate that Random Forest with high performance across all metrics was a robust classifier for predicting pendency in our dataset.

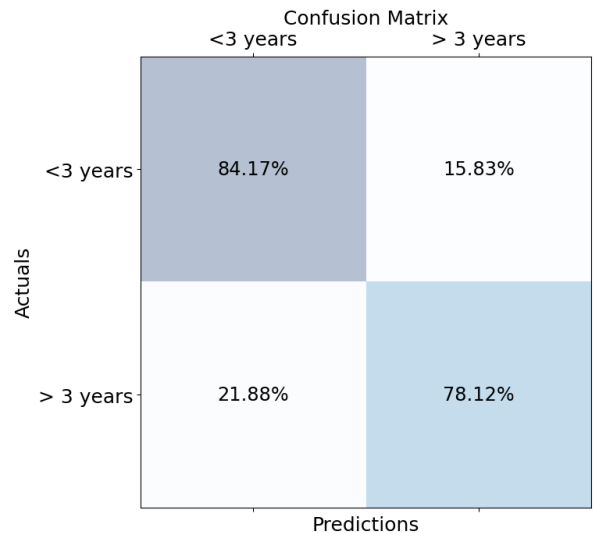

**Fig. 3.** Confusion Matrix – Custom

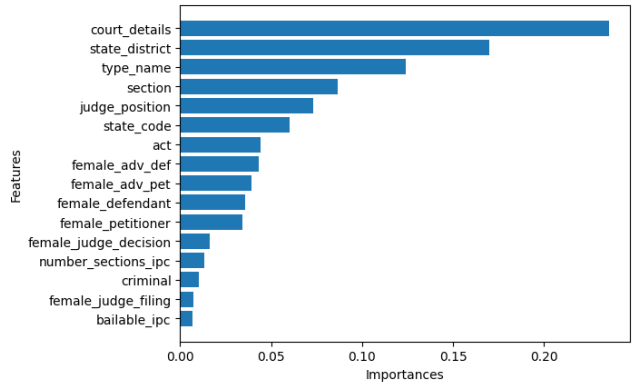

**Fig. 4.** Feature Importance (Custom Model)

**Table 4.** Comparative Results (Custom Model)

|  | **Decision Tree** | **Bagging** | **XGBoost** | **Random Forest** |
|---|---|---|---|---|
| **Accuracy** | 0.74 | 0.81 | 0.81 | 0.82 |
| **PR AUC** | 0.33 | 0.39 | 0.39 | 0.40 |
| **ROC AUC** | 0.72 | 0.81 | 0.80 | 0.81 |
| **Log loss** | 0.52 | 0.80 | 0.41 | 0.57 |
| **F1 score** | 0.73 | 0.81 | 0.81 | 0.82 |
| **Precision** | 0.74 | 0.81 | 0.81 | 0.82 |
| **Recall** | 0.74 | 0.81 | 0.81 | 0.82 |

## 5. Discussion of Key Results

The results demonstrate the feasibility of predicting delays in lower courts using available data during the early stages of the cases. The AutoML results indicate a high level of accuracy in predicting both extreme cases, specifically those with durations of less than one year and greater than ten years. Furthermore, employing a random forest binary classification model yielded comparable and satisfactory values for accuracy, F1 score, precision, and recall metrics. This model exhibited robust performance in predicting cases with delays ranging from less than three years to more than three years. The modelling revealed a feature matrix, which highlights the key factors that contribute to delays in court proceedings. Notably, the type of court, characteristics of the case, and geographical location were found to have a significant influence on the occurrence of delays (see Fig. 4). Additionally, the subject matter, such as the specific Act involved, and the complexity of the criminal code sections were also identified as factors contributing to delays. Moreover, the participation of women in the justice administration, including roles as judges, advocates, or litigants, displayed a positive correlation with case delays (see Fig. 4). These findings align with and support existing literature on the subject, except on the correlation between the participation of women in justice administration and the delay that needs further exploration.

### 5.1 State, district, court type and judge position

India has a hierarchical but unified judiciary. The judiciary is arranged into three tiers, namely the Union Judiciary (the Supreme Court), the State Judiciary (High Courts), and Subordinate Courts [29]. Administration justice at the high court level and below is supported primarily by the state governments. Similarly, the population of states, territorial limits of courts, and their operational efficiency vary significantly from one state to another [30]. There are demographic and infrastructural divergences that have a bearing on the pendency. The volume, average length of pendency and case clearance rate varies significantly among different states [31]. Even within a state, there are significant divergences in pendency [32]. Likewise, there is notable divergence in the volume of the type of cases (civil and criminal) among the states [32]. A strong correlation between state, court, and judge types and pendency, as shown in Fig. 2, is consistent with the literature [31 & 33].

**5.2 Act type, case types (criminal), number of sections and bailable offences**

Compared to petty (bailable) criminal cases, summons criminal cases take a longer time to reach finality. 90.09% of the pending criminal cases are summons cases [31]. Additionally, the number of charges under the Indian Penal Code, case and Act types are positively correlated with the length of the pendency. The stages at which cases experience delays vary, with the severity of charges and the number of sections involved playing a role in the length of pendency. For example, according to the Daksha Report, 31% of cases were pending at the "appearance" stage, 7.2% at the "framing of charges" stage, 36% at the "evidence" stage, 17.9% at the "arguments" stage, and 8% at the orders stage [34].

The severity of charges in non-bailable offences suggests that they tend to remain pending in the judicial system for a longer duration compared to bailable offences. Delays are more pronounced in session cases, which involve serious criminal offences, compared to summons cases. Furthermore, the number of charges under the Indian Penal Code, as well as the type of case and Act, positively correlates with the length of pendency. Therefore, the case-clearing rate is influenced by factors such as the type of Act, case types, the number of sections involved, and bailable offences. The modelling rightly underlines the correlation between these features.

**5.3 Role of Gender in Pendency**

In India, the overall numbers of women judges in lower courts remain low; however, in recent decades, higher proportions of women have been appointed to the lower judiciary [35]. Despite the presence of a considerable number of women judges in the judiciary, little empirical inquiry has been made into whether their experiences of judging and judicial work differ and whether they face distinct challenges that affect their optimal performance [36]. This study points out that there might be some considerable occupational challenges that women judge face that hinder their performance. Among the features that impact the duration or outcome of legal cases, six relate to the participation of women in various roles. Out of these six features, two were related to female judges. This study also indicates that the judicial administration may not be adequately responsive to and accommodating of women, as the findings reveal that court cases are more likely to experience delays when one of the parties is female, when female lawyers represent the parties, or when a female judge presides over the case. These results are somewhat unexpected, but not surprising as the gender (decisional) bias and inadequate representation of women at the bar and on the bench are well established [37]. However, the findings of this research call into question the equal application of procedural law and court processes. There is a need for further empirical investigation to test the findings of this study.

## 6. Limitations & Future work

The current modelling of pendency in Indian courts incorporates some factors that are known to contribute to delays, but there is significant room for improvement by bringing in more factors like the judge-specific data. The processing of some high cardinal categorical variables used label encoders & one-hot-encoding for encoding, the processing of which can be improved by using Hashing Encoder [38]. The model can be made more robust by including data from 2011 onwards which is available in the dataset and further fine-tuning of the models can be implemented for improving accuracy and performance.

A predictive application can also be developed which can provide this information readily to the various stakeholders involved. The absence of data on certain features such as interlocutory applications, inter-court appeals, and stays of proceedings by higher courts limits the ability to examine their impact on case duration. To address this, a more comprehensive data set can be further constructed to increase the accuracy and insights of the predictive model. In particular, more detailed data on the categorisation at the court and district levels is required to forecast pendency more effectively. Although the existing data set provides information on courts, districts and states, unique codes are needed to locate specific courts or districts within a state, which would improve the ability to predict the duration of the case.